\documentclass[runningheads]{llncs}

\usepackage{accv}

\usepackage{accvabbrv}

\usepackage{graphicx}
\usepackage{booktabs}
\usepackage{multirow}

\usepackage[accsupp]{axessibility}  

\usepackage{hyperref}

\usepackage{orcidlink}

\begin{document}

\title{RootDown: Deploying and Benchmarking Models for Beet-Root Detection on Edge Devices}
\titlerunning{Beet-Root Detection on Edge Devices}

\author{Gerardus Croonen\orcidlink{0009-0001-4470-6705} \and
Andreas Trondl\orcidlink{0009-0006-5748-3607} \and
Julia Simon\orcidlink{0000-0003-0243-9706} \and
Daniel Steininger\orcidlink{0000-0003-3810-1803}}
\authorrunning{G.~Croonen et al.}

\institute{AIT Austrian Institute of Technology, Giefinggasse 4, 1210 Vienna, Austria\\
Center for Vision, Automation \& Control\\
\email{\tt\small gerardus.croonen@ait.ac.at}}

\maketitle

\begin{abstract}
Quality inspection of sugar beets after harvest and storage is critical to preserve sugar content and ensure optimal processing efficiency. The transfer of harvested beets to field clamps or trucks provides an opportunity to visually inspect individual beets automatically on loader conveyors. However, this is a challenging task as the large number of beets causes partial overlap and occlusion, complicating both annotation and detection. In this work, we propose an end-to-end image-based pipeline for detecting partially occluded sugar beets in loading scenarios as a first step toward automated quality inspection. We introduce a dedicated benchmark and present an automatic annotation strategy that leverages the SAM 3 foundation model to generate bounding boxes. We compare models trained on manually annotated and automatically generated labels. Additionally, we investigate deployment considerations by benchmarking multiple inference configurations, including execution on an edge device, under realistic operational constraints regarding precision, inference time and power consumption. Our results show that, despite substantial occlusion and inherent annotation ambiguity, foundation-model-based labels can substitute for a substantial share of manual annotation, trading labeling effort for data volume at a modest accuracy cost. Furthermore, we demonstrate that a Raspberry Pi 5 equipped with a Hailo accelerator enables high-performance, low-power inference, making it a practical solution for mobile outdoor deployment.
\end{abstract}

\section{Introduction}
\begin{figure}
    \centering
    \includegraphics[width=1.0\columnwidth]{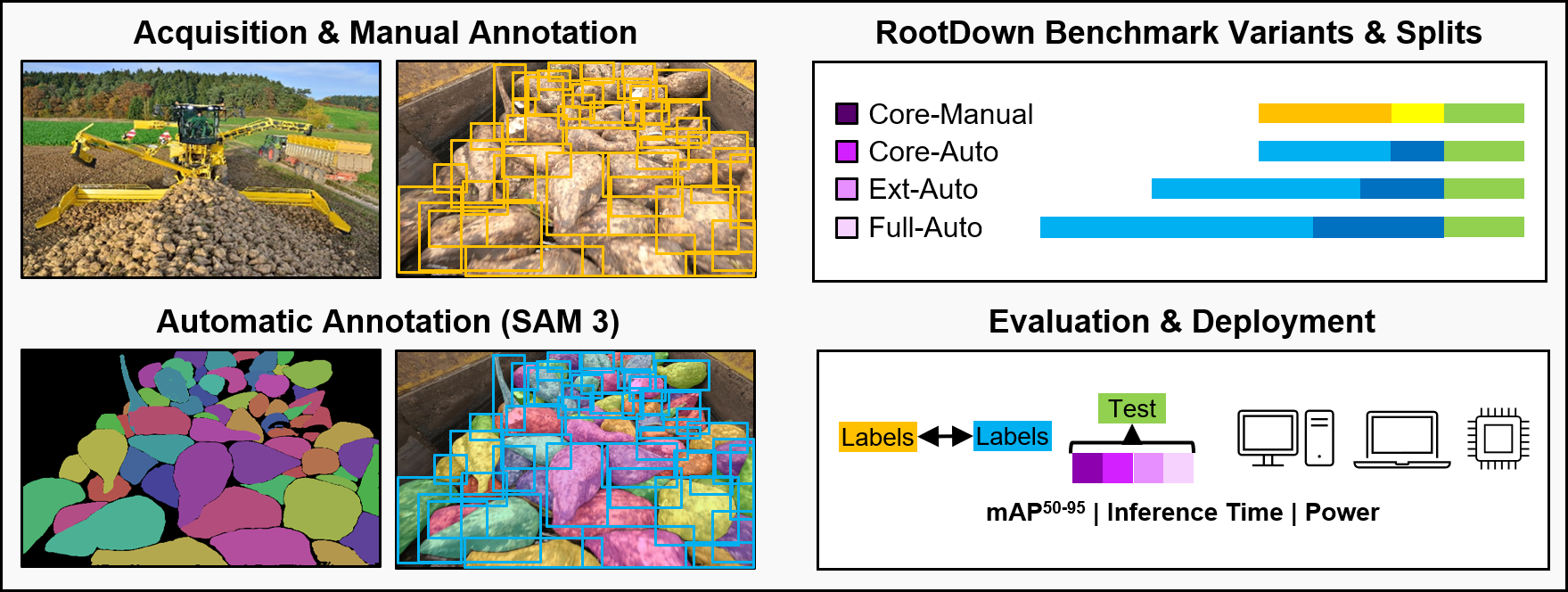}
    \caption{Overview of the proposed pipeline and evaluation protocol, spanning data acquisition, annotation, training, evaluation and deployment on edge hardware. Multiple benchmark variants (pink) are defined using manually and automatically annotated data to assess both automatic annotation quality and deployment characteristics (orange, yellow and green: manually annotated training, validation and test splits; light and dark blue: automatically annotated training and validation splits).}
    \label{fig:approach_overview}
\end{figure}

The availability of sugar beets during harvest far exceeds the processing capacity of sugar factories, resulting in most beets being stored in field clamps, which typically consist of several hundred thousand beets. After storage periods of up to 90 days, the clamps are transferred to sugar factories by truck, where a human estimator assesses beet quality by visual inspection, focusing on soil adhesion, topping quality and the presence of mold, rot and vegetation. Detached soil settling at the bottom of truck beds during transfer reduces the effectiveness of visual inspection, which is limited to the top layer of beets. Internal evaluations at sugar factories show that human estimators mostly base quality estimates on overall appearance and generally underestimate total soil tare. Time constraints limit the inspection of individual beets to a small number of samples. The use of an automated image-based visual inspection system during transfer can provide many advantages over manual inspection, as individual beets can be observed directly after rudimentary in-loader cleaning before soil accumulation occurs.

A vital first step towards automated visual inspection of sugar beets is their robust detection and separation. Therefore, the aim of this work is to develop, evaluate and deploy a real-time image-based method for detecting beets with moderate to high visual overlap on conveyors. We present an end-to-end pipeline that integrates data acquisition, automatic annotation, training, evaluation and deployment across various hardware platforms, and characterizes performance in terms of inference time, precision and power consumption (\cref{fig:approach_overview}), providing the following key contributions:

\begin{enumerate}
    \item
        We introduce \textit{RootDown}, the first \textbf{benchmark} for partially overlapping sugar beets captured on conveyors during realistic loading operations, and propose a scalable annotation strategy that integrates a foundation model to substantially reduce manual labeling effort.\footnote{Benchmark, code and models: \href{https://github.com/semanticsugarbeets/rootdown}{https://github.com/semanticsugarbeets/rootdown}}        
    \item
        We \textbf{quantify} the impact of manually curated and automatically generated annotation sources, recording conditions and camera viewpoints on detection performance through systematic experimentation.
    \item
        We \textbf{deploy} an end-to-end system for sugar-beet detection, systematically measuring detection accuracy, inference time and power consumption across a diverse range of embedded and general-purpose hardware platforms.
\end{enumerate}
\section{Related Work}
Weather-based decision-support tools predict soil moisture and stickiness to optimize harvest timing \cite{GaHa19} and reduce soil tare. Operational optimizations of beet lifters, such as inducing beet motion, reduce soil adhesion in wet clay soils \cite{VeKo02}. In-loader cleaning mechanisms maximize soil removal through mechanical impact while limiting beet damage \cite{DePe01}. Camera-based quality assessment of agricultural produce has been studied for decades \cite{BrSu02, BaSh14, RaBa23}, yet applications to root vegetables such as sugar beets remain limited. A camera-based yield-estimation system based on ellipsoid reconstruction of potatoes is proposed in \cite{HoMo03}. The \textit{SemanticSugarBeets} dataset \cite{CrTr25} focuses on segmentation but is limited to non-overlapping beet scenarios. Early automated inspection systems employed classical machine-vision pipelines that combined segmentation, handcrafted features and classifiers such as k-NN, PCA, LDA, SVM or RVM \cite{HeRa22, ZhHu15, AsSh18, PaPr19}. While these approaches showed promising results, they were typically evaluated under unrealistic conditions, including controlled backgrounds and complete object isolation. 

CNN-based methods have substantially improved detection and segmentation performance in agricultural applications such as fruit counting, nutrient deficiency detection \cite{YiKr20}, defect segmentation, yield estimation and seed coating inspection \cite{BeSa25}. Modern detectors such as Faster R-CNN \cite{SaGe16, BaUn17}, YOLO variants and transformer-based architectures perform well in cluttered environments but require large, carefully annotated datasets. Conveyor-based object detection has been studied in industrial domains such as mining, food processing, recycling and retail. Produce inspection systems for tomatoes, oranges, mangoes and lentils have been reported under low-throughput conditions with fixed lighting and no object overlap \cite{TrVu24, NaAs19, RaiH26, YaSa25}. A YOLO-based foreign-object detector for cut vegetables is presented in \cite{KuAr25}. Geometric reconstruction and size estimation of aggregate materials on conveyors using morphological operators is proposed in \cite{KoMi17}. Occlusion-aware detection has been addressed through part-based models \cite{AgAw04, FeGi09, ZhDo14}, attention mechanisms \cite{LiZe18, ZhZh23, HeWa22} and augmentation strategies \cite{ZhZh20, KaKh21, FoVe19}. Explicit overlap-modeling approaches include occlusion-aware human pose estimation \cite{RaGa15} and UOAIS \cite{BaLe22}, which models visible, amodal and occlusion masks to infer object ordering. Such methods are more challenging to apply to natural objects such as sugar beets because of their high intra-class variability and lack of a stable semantic part structure. 

Foundation vision models such as Segment Anything 3 (SAM 3) \cite{CaGu25} and CLIPSeg \cite{LuEc22} demonstrate strong zero-shot segmentation capabilities but are unsuitable for real-time edge deployment due to their high computational demands. Recent work applies such models to annotation bootstrapping and domain adaptation, accelerating dataset creation; prompt-ensemble fusion further improves mask quality \cite{LiKe24, MuFu24, ZhZh25}. Although (semi-)automatic annotation, interactive segmentation and active learning reduce labeling effort, their performance depends on strong initial proposals and degrades under heavy clutter and occlusion \cite{PaZi18, SiVa20}. Despite growing interest in agricultural vision systems, the detection of partially overlapping natural objects, particularly root crops in operational settings, remains underexplored. Existing methods are rarely evaluated on energy- and compute-constrained platforms. We address this gap through a dedicated benchmark, a semi-automatic annotation workflow and an in-depth analysis of model training and deployment under realistic constraints.
\section{RootDown Benchmark}
As no publicly available image dataset exists for sugar-beet detection on conveyors, we acquire image data under operational conditions across various camera positions, loading locations and lighting conditions. Using efficient manual annotation strategies and, for comparison, a fully automatic annotation method, we create a dedicated benchmark for sugar-beet detection on conveyors.

\begin{figure}
    \centering
    \includegraphics[width=0.75\columnwidth]{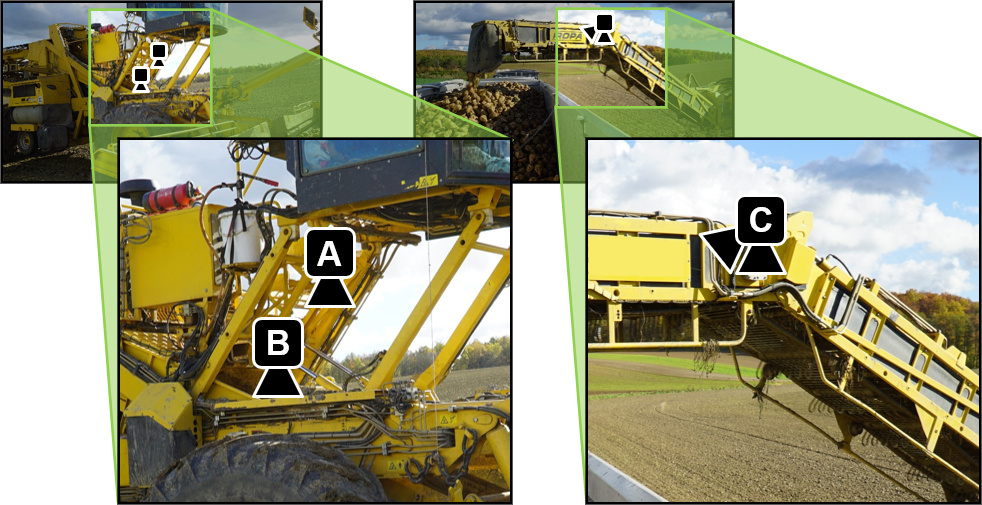}
    \caption{Camera positions and viewpoints used in three scenarios: \textit{Front} (A/B, top view), \textit{Side} (C, side view) and \textit{Back} (C, top view).}
    \label{fig:camera_mount_points}
\end{figure}

\subsection{Image Acquisition, Sampling and Meta Annotation}
Image acquisition is performed on the conveyors of a commercial beet loader during unscripted loading operations. Three GoPro Hero 9 cameras are mounted at selected positions on the front and back conveyors (\cref{fig:camera_mount_points}) to capture multiple viewpoints for subsequent comparative evaluation. Recordings are conducted over three days and 52 loading sessions using both regular (30 fps) and time-lapse (1 fps) camera modes. After recording, sequences are sampled uniformly at 1 fps and trimmed to exclude periods during which the conveyor is stationary. To complement the operational recordings, high-resolution images of soil-covered beets are captured on the ground using a Sony Alpha S camera. Finally, 597 images are randomly selected from the extracted material to balance the viewpoint distribution and minimize redundancy. The selected images are resized to a height of 1080 pixels while preserving the aspect ratio.

\subsubsection{Meta Parameters.} 
\begin{figure}
    \centering
    \includegraphics[width=1.0\columnwidth]{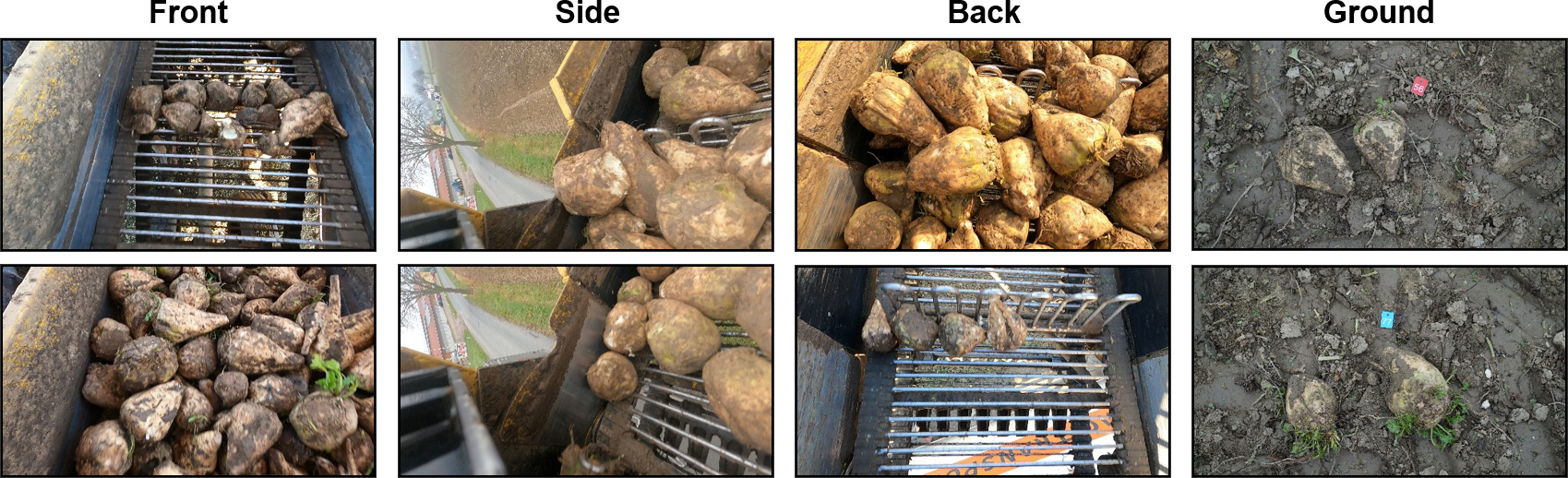}
    \caption{Representative image samples illustrating the appearance variability across the benchmark scenarios \textit{Front}, \textit{Side}, \textit{Back} and \textit{Ground}.}    
    \label{fig:overview_scenarios}
\end{figure}

To facilitate the analysis of benchmark characteristics and model performance under varying conditions, all images are manually annotated with the following attributes:

\begin{itemize}
    \item 
        \textbf{Scenario} denotes the camera viewpoint, with \textit{Front} referring to camera mount points on the loader's front conveyor, \textit{Back} and \textit{Side} referring to mount points on the rear lifting conveyor, and \textit{Ground} denoting manually captured images acquired in the field. \cref{fig:overview_scenarios} shows representative recordings for each acquisition scenario. 
    \item 
        \textbf{Lighting} defines the natural lighting conditions as \textit{Sunny}, \textit{Diffuse} or a combination of both (\textit{Mixed}).
    \item 
        \textbf{Location} differentiates between four recording sites.        
\end{itemize}

\begin{figure}
    \centering
    \includegraphics[width=0.65\columnwidth]{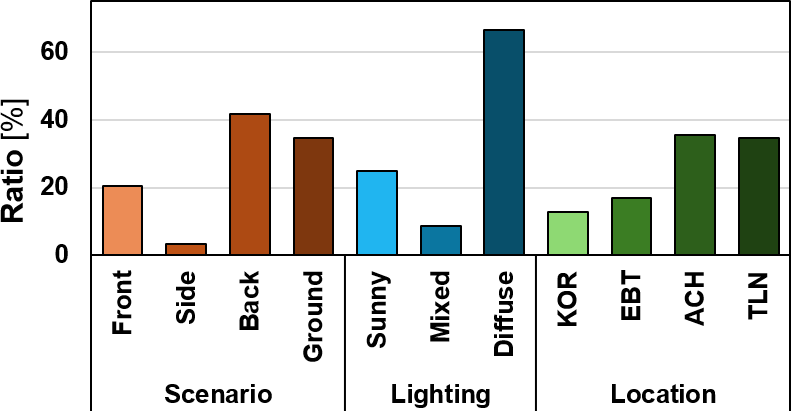}
    \caption{Relative distributions of annotated benchmark samples across acquisition scenarios, lighting conditions and recording locations.}
    \label{fig:dataset_statistics}
\end{figure}

The distributions of the annotated meta-parameters, summarized in \cref{fig:dataset_statistics}, indicate that the benchmark covers a realistic range of lighting conditions. Diffuse lighting is common due to the frequent occurrence of foggy and overcast weather during the autumn and winter beet harvest seasons. Parameters such as \textit{Scenario} and \textit{Location} are fairly evenly distributed, with the exception of the rare \textit{Side} scenario, which results from severe vibrations causing the camera to rotate from the \textit{Back} viewpoint into a side-facing orientation.
 
\subsection{Manual and Automatic Annotation}
Manual annotation is performed using CVAT \cite{Cv26} by assigning bounding boxes to individual beets in 312 randomly selected images, resulting in a total of 6,336 annotated instances and requiring approximately 25 hours of annotation effort.

\begin{figure}
    \centering
    \includegraphics[width=1.0\columnwidth]{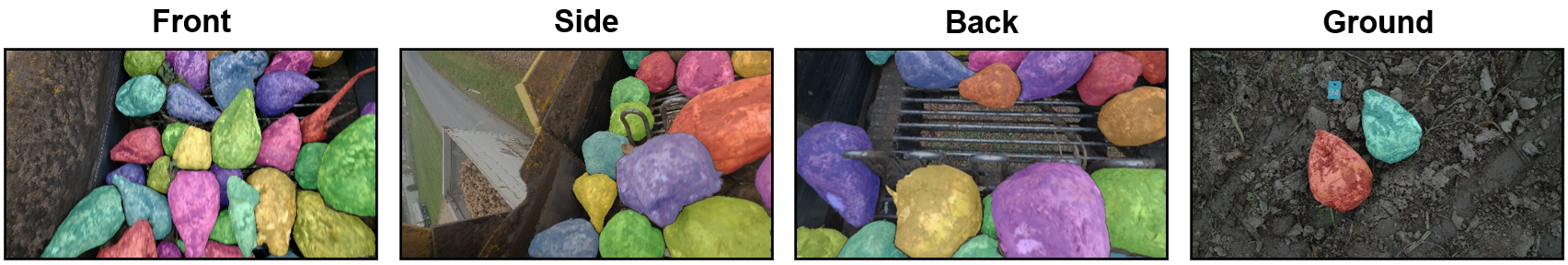}
    \caption{Representative instance masks generated by SAM 3 using the text prompt \textit{vegetable}, overlaid on corresponding sample images for each scenario.}
    \label{fig:sam3_instances}
\end{figure}

The SAM 3 foundation model \cite{CaGu25} is employed to automatically extract object masks from all 597 images using text prompts. Counterintuitively, neither highly specific terms such as \textit{sugar beet}, \textit{beet root}, \textit{beet} or \textit{root}, nor generic descriptions such as \textit{round objects} or \textit{balls} reliably produced accurate masks. In contrast, the broader yet domain-appropriate prompt \textit{vegetable} consistently yielded the most accurate segments across all scenarios. As illustrated in \cref{fig:sam3_instances}, the generated masks generally capture object contours well, although they may contain interior holes, which are inconsequential for bounding-box derivation. Bounding boxes are derived directly from the mask contours obtained using the prompt \textit{vegetable} for all 597 images. SAM 3 recovers approximately 90\% of manually identified instances (5,709 vs.\ 6,336). To assess agreement between manual and automatic annotations, we compute mAP\textsuperscript{50--95} using \textit{pycocotools} \cite{LiMa14}, treating the manual annotations as ground truth and the SAM 3 outputs as detections. This yields a score of 46.1\%, suggesting that, in our setting, SAM 3-generated bounding boxes are suitable for training object-detection models. Beyond annotation quality, generating annotations for the entire benchmark requires only 20 minutes on an NVIDIA RTX 3090 GPU, highlighting the efficiency of this approach.

\section{Methodology}
Object detection constitutes the first stage of the proposed real-time quality-inspection pipeline due to its computational efficiency and versatility. Accurate bounding boxes provide the foundation for downstream tasks such as object counting, region-based classification and soil-adhesion estimation from semantically segmented individual beets or aggregated per-beet statistics. As these tasks depend on reliable object localization, object detection serves as a natural entry point for the pipeline. The present work focuses on this detection stage, while downstream components are left for future development.

\subsection{Benchmark Ablation}
To assess the influence of automatic annotation on model performance, we construct several benchmark variants. Our baseline variant, \textit{Core-Manual}, consists exclusively of manually annotated data and uses a 50:20:30 split for training, validation and test sets, respectively. Analogously, \textit{Core-Auto} uses the same image split but replaces the manual annotations in the training and validation sets with the corresponding automatically generated labels. To investigate the effect of additional automatically annotated data, we extend \textit{Core-Auto} with extra SAM 3-annotated images, resulting in the \textit{Ext-Auto} variant. Finally, we define \textit{Full-Auto}, which uses all automatically annotated images available in the benchmark. An overview of all benchmark variants, including image counts, split ratios and annotation types, is provided in \cref{tab:subdatasets}.

\begin{table}
    \caption{Benchmark variants used for experimentation, all based on the same image split. To ensure fair comparison, the test-set images and annotations are identical across all variants and consist exclusively of manually annotated data.}    
    \label{tab:subdatasets}
    \centering
    \begin{tabular}{@{}l@{\quad}lll@{\quad}rrr@{\quad}rrrr@{}}
        \toprule
        &
        \multicolumn{3}{c@{\quad}}{\textbf{Annotation}} &
        \multicolumn{3}{c}{\textbf{Ratio}} &
        \multicolumn{4}{c}{\textbf{Images}} \\
        \textbf{Benchmark} &
        Train & Val & Test &
        Train & Val & Test &
        Train & Val & Test & Total \\    
        \midrule
        \textit{Core‑Manual}
            & Man  & Man  & Man
            & 50   & 20   & 30
            & 156  & 62   & 94
            & 312 \\        
        \textit{Core‑Auto}
            & Auto & Auto & Man
            & 50   & 20   & 30
            & 156  & 62   & 94
            & 312 \\       
        \textit{Ext‑Auto}
            & Auto & Auto & Man
            & 55   & 24   & 21
            & 250  & 110  & 94
            & 454 \\       
        \textit{Full‑Auto}
            & Auto & Auto & Man
            & 59   & 26   & 16
            & 350  & 153  & 94
            & 597 \\
        \bottomrule
    \end{tabular}
\end{table}

\subsection{Hardware Setup}
\begin{figure}
    \centering
    \includegraphics[width=0.4\columnwidth]{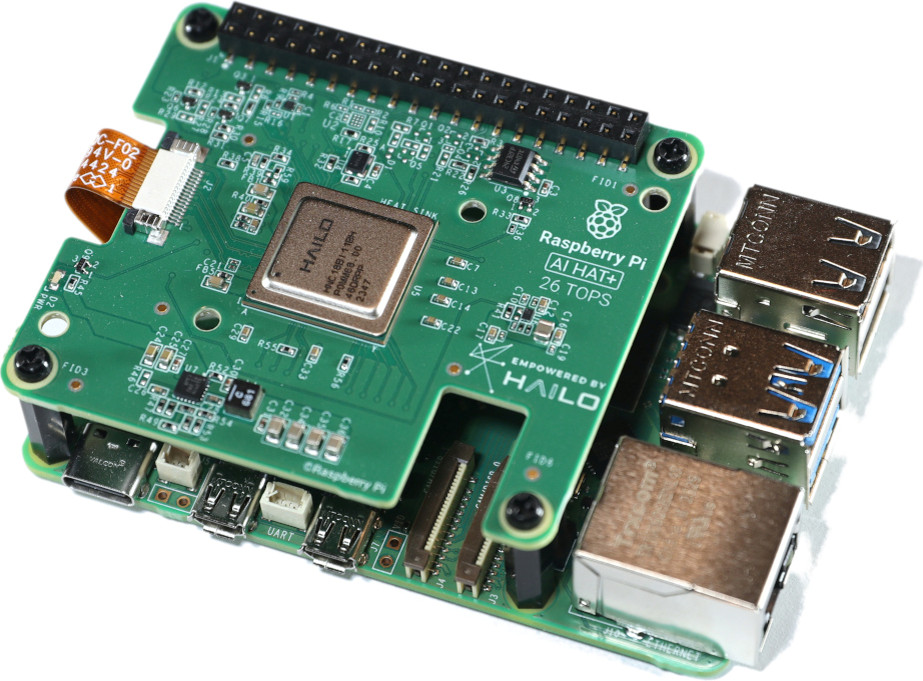}
    \caption{Raspberry Pi 5 single-board computer with a Hailo-8 AI accelerator extension board mounted on top.}
    \label{fig:rpi5_with_npu_hat}
\end{figure}

To evaluate model performance across a range of computational scenarios, we select hardware platforms with varying levels of processing power, energy consumption and mobility. Our selection, summarized in \cref{tab:hardware_comparison}, includes laptop and workstation CPUs, a workstation GPU, and a single-board computer (SBC), both with and without a neural processing unit (NPU). For the SBC platform, we select the Raspberry Pi 5, which offers native support for Hailo AI edge accelerators (\cref{fig:rpi5_with_npu_hat}). In particular, the Hailo-8L and Hailo-8 modules provide up to 13 and 26 TOPS of compute, respectively, combined with low power consumption. The Hailo toolchain converts pretrained models into a dedicated binary format (HEF, Hailo Executable File), which can be executed directly from a Python environment. Version 4.20 of both the Hailo PCIe driver and the associated tool suite is used for all experiments.

\subsection{Model Training}
The model architecture used throughout the experiments must be compatible with all deployment targets, with the edge platform imposing the most restrictive constraints. We therefore employ the YOLO11 architecture \cite{JoCh24} for all experiments, as YOLO-based detectors are well supported and achieve competitive accuracy on Hailo hardware \cite{Ha26}. The Ultralytics YOLO11 framework \cite{JoCh24} is used for training, as it provides stable model variants, an established training pipeline and seamless model export. Pretrained YOLO11-Large models are fine-tuned for 100 epochs using a batch size of eight and input resolutions of $640^2$ and $1024^2$ pixels. Unless otherwise stated, all hyperparameters are kept at their default values. Benchmark annotations and model predictions are stored as COCO-format JSON files for evaluation. The precision metrics reported in this work are computed using the standard \textit{pycocotools} implementation.

\begin{table}
    \caption{Overview of the evaluated hardware platforms, showing INT8-equivalent TOPS (trillion operations per second) and TDP (thermal design power) in watts, sorted by TOPS per watt. Values shown in \textit{italics} denote estimates derived from vendor-reported throughput figures.}    
    \label{tab:hardware_comparison}
    \centering
    \begin{tabular}{@{}l@{\quad}ll@{\quad}rr@{}}
        \toprule
        \textbf{Platform} & \textbf{Type} & \textbf{Design} & \textbf{TOPS} & \textbf{TDP} \\
        \midrule
        Intel W-2125 & CPU & x86-64 & \textit{1.2} & 120 \\
        Raspberry Pi 5 & CPU & ARM64 & \textit{0.6} & 7 \\
        AMD 7430U & CPU & x86-64 & \textit{1.5} & 15 \\
        AMD 9950X3D & CPU & x86-64 & \textit{20.0} & 170 \\
        NVIDIA RTX 3090 & GPU & CUDA & \textit{568.0} & 350 \\
        \midrule
        Hailo-8 on RPi 5 & NPU & Custom & 26.0 & 6 \\
        Hailo-8L on RPi 5 & NPU & Custom & 13.0 & 3 \\
        \bottomrule
    \end{tabular}
\end{table}

\subsection{Model Deployment}
The Python-based Ultralytics training framework supports multiple export formats for x86-64, ARM and CUDA-based platforms. We select ONNX due to its broad compatibility with CPU- and GPU-based systems, supporting both FP16 and FP32 inference. OpenVINO is additionally evaluated because it provides optimized INT8 quantization specifically targeting CPU deployment. TensorRT is excluded, as it is optimized exclusively for CUDA-enabled GPUs, preventing a direct comparison across all target platforms. INT8-compressed models are obtained through post-training quantization using calibration data from the training set to determine suitable scaling parameters and biases. Deployment on Hailo devices relies on the Hailo Dataflow Compiler (DFC). The parsed model’s non-maximum suppression (NMS) block is replaced by a hardware-optimized equivalent, and the network is quantized using a calibration set of randomly selected training images. Models are compiled separately for the Hailo-8 and Hailo-8L to evaluate both accelerators independently.

\subsection{Real-Time Optimization and Power Constraints}
Power consumption is a critical consideration for edge devices, particularly in mobile and outdoor applications. Total power draw can be decomposed into idle power, consumed regardless of workload, and marginal power, defined as the additional power required to perform useful computation. In practice, however, measuring idle and marginal power separately is technically challenging and of limited practical relevance. For real-world applications, the key metric is the total energy consumed per unit of useful work while processing a continuous camera stream at a fixed frame rate. This measure directly determines system-level design constraints, such as the battery capacity required to sustain operation for a specified duration.

\subsubsection{Runtime Measurements.}
To characterize runtime behavior, we measure inference times across all evaluated devices using trained YOLO11 models spanning a range of complexities from Nano to X-Large. To ensure consistent and reproducible measurements, inference time is measured for each device--model combination over 500 frames using randomized dummy input tensors. An additional 100 frames are processed as a warm-up phase before timing begins, and the reported values are averaged across all measured frames.

\subsubsection{Power Requirements.}
To estimate power and energy requirements under realistic conditions, we measure total power consumption and the number of processed images over time on a Raspberry Pi 5 single-board computer paired with a Hailo-8 accelerator. From these measurements, we derive the average energy cost per inference. Idle power draw is recorded separately to distinguish baseline consumption from workload-dependent power usage.
\section{Results}
This section summarizes our results on detection accuracy, inference time and power consumption across benchmark variants, models and hardware platforms.

\begin{figure}[t]
    \centering
    \includegraphics[width=1.0\columnwidth]{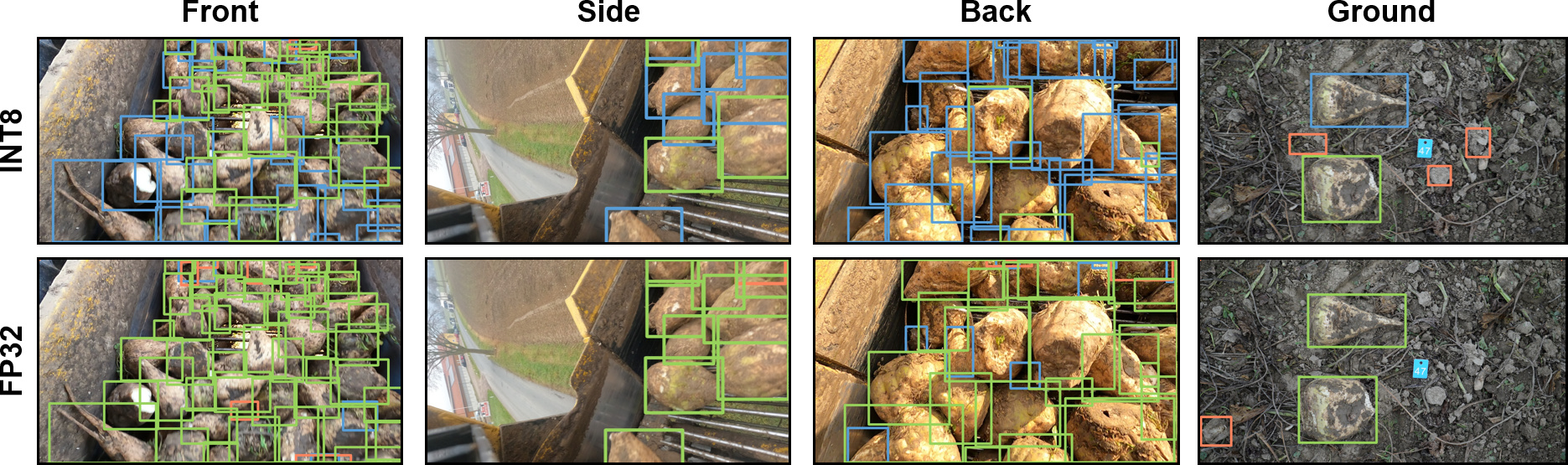}
    \caption{Representative detection results on test images for a YOLO11-Large model with an input resolution of $640^2$ pixels deployed on a Raspberry Pi 5 using OpenVINO in FP32 and INT8 precision (green: correct detections, red: incorrect detections, blue: missed instances).}
    \label{fig:qualitative_results}
\end{figure}

\subsection{Detection Accuracy}

\begin{table}[b]
    \caption{Detection performance (\% mAP\textsuperscript{50--95}) of YOLO11-Large models at an input resolution of $640^2$ pixels trained on the \textit{Core-Manual} benchmark variant across different deployment platforms, model formats and numerical precisions. An \textit{x} indicates unsupported combinations, a colon denotes identical values across a column. Hailo accelerators were evaluated exclusively as Raspberry Pi 5 HATs.}
    \label{tab:mAP_per_platform}
    \centering
    \begin{tabular}{@{}lc@{\quad}c@{\hspace{0.5em}}c@{\quad}c@{\hspace{0.5em}}c@{\hspace{0.5em}}c@{\quad}c@{\hspace{0.5em}}c@{}}
        \toprule
        & \multicolumn{1}{c@{\quad}}{\textbf{PyTorch}} & \multicolumn{2}{c@{\quad}}{\textbf{ONNX}} & \multicolumn{3}{c@{\quad}}{\textbf{OpenVINO}} & \multicolumn{2}{c}{\textbf{Hailo}} \\
        \textbf{Platform} & FP32 & FP32 & FP16 & FP32 & FP16 & INT8 & 8 & 8L \\
        \midrule
        Raspberry Pi 5 & : & : & : & 56.0 & 56.0 & 23.3 & 58.7 & 56.2 \\
        Intel W-2125 & : & : & : & 56.0 & 56.0 & 56.3 & x & x \\
        AMD 7430U & 56.9 & 56.0 & 56.0 & 56.0 & 56.0 & 56.2 & x & x \\
        AMD 9950X3D & : & : & : & 55.8 & 55.9 & 56.2 & x & x \\
        NVIDIA RTX 3090 & : & : & : & x & x & x & x & x \\
        \bottomrule
    \end{tabular}
\end{table}

The detection performance of YOLO11-Large models across all devices, deployment frameworks and precisions for the \textit{Core-Manual} benchmark variant is summarized in \cref{tab:mAP_per_platform}. As expected, models with identical formats and precisions achieve nearly identical accuracy across platforms, with minor deviations for OpenVINO FP32 and FP16. INT8 quantization generally yields slightly higher accuracy, except for OpenVINO on the Raspberry Pi 5. Despite extensive efforts to improve performance, including exports from an x86-64 host, re-exporting on the SBC, varying opsets and adjusting calibration sets, accuracy remained low, potentially due to limitations of ARM CPUs when accumulating 8-bit integer operations. \cref{fig:qualitative_results} shows a qualitative comparison of results.

On the Hailo-8L and Hailo-8 accelerators, deployed models achieve mAP\textsuperscript{50--95} scores of 56.2\% and 58.7\%, respectively. The higher accuracy of Hailo-8 is likely attributable to improved calibration and weight optimization, as well as differences in the hardware-optimized NMS implementation. The slightly lower accuracy of Hailo-8L is consistent with its more constrained resources, which affect quantization and compilation.

\begin{figure}
    \centering
    \includegraphics[width=0.65\columnwidth]{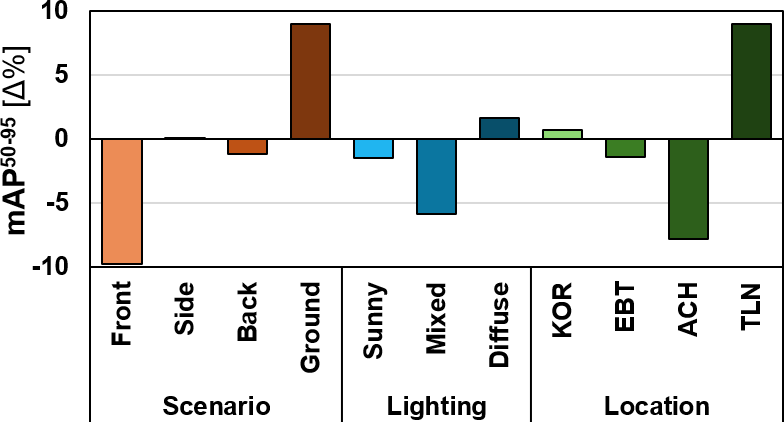}
    \caption{Impact of meta-parameters on detection performance for a YOLO11-Large FP32 PyTorch model trained on the \textit{Core-Manual} benchmark variant. Deltas are computed from image-averaged mAP\textsuperscript{50--95} and given in percentage points.}
    \label{fig:performance_impact}
\end{figure}

Using the PyTorch FP32 model, we further assess the influence of scenario, lighting and recording location on detection performance, as shown in \cref{fig:performance_impact}. Detection accuracy is lowest for the \textit{Front} scenario, likely due to shorter object distances and increased motion blur compared to the \textit{Back} scenario. Conversely, performance is highest for the \textit{Ground} scenario, in which non-overlapping beets are photographed against soil backgrounds. The model performs best under diffuse illumination, while mixed lighting reduces accuracy, reflecting the distribution of training samples. Differences across recording locations are also observable. However, due to correlations with lighting conditions, these effects cannot be attributed to location alone.

Detection results for YOLO11-Large models trained on different benchmark variants at two input resolutions are reported in \cref{tab:mAP_per_dataset_variant}. The results show a clear performance gain with increasing amounts of automatically generated training data. At $640^2$ and $1024^2$ pixels, the \textit{Core-Auto} variant achieves 42.7\% and 49.5\% mAP\textsuperscript{50--95}, respectively. Extending the automatically annotated benchmark further improves performance, with \textit{Ext-Auto} and \textit{Full-Auto} achieving progressively higher scores, reaching 50.6\% and 52.6\% mAP\textsuperscript{50--95} at the higher resolution.

\begin{table}
    \caption{Detection performance (\% mAP\textsuperscript{50--95}) of YOLO11-Large models across all benchmark variants at input resolutions of $640^2$ and $1024^2$ pixels.}
    \label{tab:mAP_per_dataset_variant}
    \centering
    \begin{tabular}{@{}l@{\quad}c@{\hspace{0.5em}}c@{}}
        \toprule
        \textbf{Benchmark} & $640^2$ & $1024^2$ \\
        \midrule
        \textit{Core-Auto} & 42.7 & \textbf{49.5} \\ 
        \textit{Ext-Auto} & 48.0 & \textbf{49.7} \\
        \textit{Full-Auto} & 50.6 & \textbf{52.6} \\
        \midrule
        \textit{Core-Manual} & \textbf{56.9} & 55.3 \\    
        \bottomrule
    \end{tabular}
\end{table}

These results indicate that, for the detection of overlapping sugar beets, automatically generated annotations can provide useful detection performance at minimal labeling effort despite their imperfections. However, models trained on the manually annotated baseline consistently outperform those trained exclusively on automatically generated annotations. Interestingly, increasing the input resolution benefits the automatically annotated benchmark variants more than the manual baseline, suggesting that higher-resolution inputs help mitigate label noise. Overall, the results support the use of foundation-model-assisted annotation for constructing larger training benchmarks, while manually annotated data remains the upper bound on performance.

\subsection{Runtime Analysis}
\cref{tab:model_ablation} reports inference times for different compute platforms and YOLO11 model sizes at an input resolution of $640^2$ pixels using FP32 precision. As expected, inference speed varies substantially across devices, with the RTX 3090 GPU delivering the highest throughput. However, the Hailo-8 on the Raspberry Pi 5 is the second-fastest configuration, being only 2.5--6.8$\times$ slower than the GPU while remaining 37.5\% (\textit{Nano}) to 53.6\% (\textit{X-Large}) faster than a modern workstation CPU such as the AMD Ryzen 9 9950X3D. The Hailo-8L trails slightly behind the Hailo-8, consistent with its reduced hardware capabilities. Inference on the Raspberry Pi 5 CPU alone is the slowest across all evaluated devices. 

\begin{table}[b]
    \caption{Inference time (ms) across compute platforms and YOLO11 model sizes (\textbf{N}ano, \textbf{S}mall, \textbf{M}edium, \textbf{L}arge and \textbf{X}-Large) at FP32 precision (where applicable) and an input resolution of $640^2$. Best results are marked in bold.}
    \label{tab:model_ablation}
    \centering
    \begin{tabular}{@{}l@{\quad}r@{\hspace{0.5em}}r@{\hspace{0.5em}}r@{\hspace{0.5em}}r@{\hspace{0.5em}}r@{}}
        \toprule
        \textbf{Platform} & \textbf{N} & \textbf{S} & \textbf{M} & \textbf{L} & \textbf{X} \\
        \midrule
        Raspberry Pi 5 & 287 & 610 & 1,563 & 1,961 & 3,704 \\
        Intel W-2125 & 95 & 149 & 250 & 435 & 667 \\
        AMD 7430U & 61 & 139 & 357 & 476 & 909 \\
        AMD 9950X3D & 16 & 36 & 74 & 91 & 179 \\
        NVIDIA RTX 3090 & \textbf{4} & \textbf{5} & \textbf{6} & \textbf{11} & \textbf{20} \\
        \midrule
        Hailo-8L on RPi 5 & 13 & 26 & 44 & 70 & 116 \\
        Hailo-8 on RPi 5 & \textbf{10} & \textbf{19} & \textbf{41} & \textbf{47} & \textbf{83} \\
        \bottomrule
    \end{tabular}
\end{table}

Detailed measurements across model formats and numerical precisions are provided in the supplementary material. Interestingly, FP16 inference on the Raspberry Pi 5 CPU is faster than INT8 inference, a counterintuitive result that is likely related to kernel-level optimizations and hardware instruction support on ARM architectures. Adding the Hailo-8 accelerator dramatically improves performance, providing a 29--45$\times$ speed-up over the CPU-only configuration. Inference times obtained with different deployment formats and numerical precisions show similar trends. ONNX reduces inference time by approximately 10\% compared to the original PyTorch models, while OpenVINO provides up to a 7.5$\times$ speed-up when moving from FP32 to INT8 on the AMD Ryzen 9 9950X3D. Overall, the observed performance is consistent with the TOPS values reported in \cref{tab:hardware_comparison}.

\subsection{Power Usage}
We evaluate system-level energy consumption on a Raspberry Pi 5 with 16 GB RAM equipped with a Hailo-8 accelerator running Raspberry Pi OS Bookworm. A Microsoft LifeCam Cinema USB camera is used for image acquisition and an external USB 3.0 storage device remains connected throughout the experiments. All nonessential peripherals, including mouse, keyboard and monitor, are disconnected prior to measurement. To measure total system power consumption, a fully charged USB power bank supplies the device while a Hailo-optimized YOLO11-Small model continuously processes live camera input at 30 fps. At the applied input resolution of $640^2$ pixels, this model variant achieves an mAP\textsuperscript{50--95} of 52.9\%. The system operates for 7 hours, 32 minutes and 33 seconds, consuming 62.6 Wh with an average power draw of 8.29 W. During this period, 825,390 images are processed, corresponding to an average energy cost of 0.0758 mWh or 272.9 mJ per frame. 

In a second experiment, idle power consumption is measured while the system remains fully connected but performs neither image capture nor inference. The power bank delivers 63.2 Wh over 18 hours, 34 minutes and 5 seconds, corresponding to an average idle power draw of 3.4 W. This baseline consumption reflects the energy requirements of the CPU, camera, NPU and storage subsystem even in the absence of useful computation. These results demonstrate that a Raspberry Pi 5 paired with a Hailo-8 accelerator can sustain continuous inference for more than 7.5 hours using a consumer-grade 20,000 mAh \mbox{(3.7 V)} USB power bank. Operating time can be extended substantially when inference is intermittent, user-triggered or combined with low-power modes, making this hardware configuration suitable for mobile and outdoor deployments with moderate energy constraints.
\section{Conclusion}
This work demonstrates that foundation-model-based annotation can substantially reduce manual labeling effort in agricultural object-detection pipelines. Although domain-specific prompts such as \textit{sugar beet} did not yield reliable segmentation, SAM 3 produced high-quality masks for overlapping sugar beets when prompted with the broader term \textit{vegetable}, recovering 90\% of instances and achieving 46.1\% mAP\textsuperscript{50--95} relative to manual annotations. These results highlight the potential of foundation models to support large-scale benchmark creation with minimal human intervention. For sugar-beet detection on conveyors, a YOLO11-Large model trained on fully manual annotations reaches 56.9\% mAP\textsuperscript{50--95} at $640^2$ and 55.3\% at $1024^2$, while models trained exclusively on automatically generated labels reach 50.6\% and 52.6\%, respectively. The remaining gaps at matched resolution must be considered in light of the approximately 25 hours required for manual bounding-box annotation, highlighting the efficiency of automated annotation at scale. Our analysis further shows that deployed-model performance depends strongly on platform-specific optimizations. On x86-64 CPUs, OpenVINO INT8 models provide the lowest inference times without sacrificing accuracy, while OpenVINO FP16 performs best on the Raspberry Pi 5 CPU. Integrating a Hailo-8 accelerator enables efficient, sustained and fully untethered operation using commodity USB power banks, providing a practical and energy-efficient edge solution for real-time quality inspection in agricultural environments.

\subsubsection{Outlook.} 
Future work should explore the scalability of automatic annotation and its combination with manual annotations, focusing on how benchmark size affects the gap between supervised and automated training. A systematic scaling study would reveal whether automatically generated labels narrow this gap and whether performance saturates due to label noise or limited data variability.  Improvements could come from confidence-based filtering, iterative self-training or prompt-ensemble strategies to reduce sensitivity to domain-specific phrasing. Semi-synthetic 2D or 3D rendered images could cover edge cases and improve robustness under challenging conditions, such as high contrast, dim or night-time lighting or wet beets. Beyond detection, integrating instance and semantic segmentation would enable estimation of soil adhesion, beet tare, damage and vegetation. Object tracking and scene-scale estimation can further enhance operational metrics. Evaluating robustness across illumination, occlusion and conveyor speeds will increase practical applicability. Finally, additional edge-optimization efforts such as model compression, energy-efficiency analysis and device connectivity will improve feasibility for sustained autonomous field operation in real-world settings.

\paragraph{Acknowledgements}
The authors would like to thank Marion Seiter, Wibke Imgenberg and Gereon Heller of the AGRANA Research and Innovation Center (ARIC) for their instrumental contributions to this research. This work was created within a research project of the Austrian Competence Center for Feed and Food Quality, Safety and Innovation (FFoQSI), which is funded by the Austrian federal ministries BMWET and BMIMI as well as the Austrian provinces Lower Austria, Upper Austria and Vienna within the scope of COMET - Competence Centers for Excellent Technologies. The program COMET is managed by the Austrian Research Promotion Agency FFG.

\bibliographystyle{splncs04}
\bibliography{references}

\newpage

\begin{center}
    \Large
    \textbf{RootDown: Deploying and Benchmarking Models for Beet-Root Detection on Edge Devices (Supplementary Material)} \\
    \vspace{2em}
\end{center}
    
\appendix
This supplementary material provides additional details and analyses that complement the benchmark description and quantitative results presented in the main paper.

\section{Extended Benchmark Statistics}
\cref{tab:dataset_statistics} presents extended benchmark statistics for the \textit{Core-Manual} benchmark variant, including sample distributions across the meta-parameters \textit{Scenario}, \textit{Lighting} and \textit{Location}.

\begin{table}
    \caption{Detailed benchmark statistics for all annotated parameters of the \textit{Core-Manual} benchmark, showing the total numbers of \textbf{Rec}ording sessions and \textbf{Samples} for each state.}
    \label{tab:dataset_statistics}
    \centering
    \resizebox{0.45\linewidth}{!}{
    \begin{tabular}{@{}l@{\quad}l@{\quad}r@{\quad}r@{}}
        \toprule
        \textbf{Parameter} & \textbf{State} & \textbf{Rec} & \textbf{Samples} \\
        \midrule
        \multirow{4}{*}{\textbf{Scenario}} & Front & 18 & 64 \\
         & Side & 2 & 10 \\
         & Back & 35 & 130 \\
         & Ground & 1 & 108 \\
         \midrule
        \multirow{3}{*}{\textbf{Lighting}} & Sunny & 21 & 78 \\
         & Mixed & 7 & 27 \\
         & Diffuse & 28 & 207 \\
         \midrule
        \multirow{4}{*}{\textbf{Location}} & KOR & 11 & 40 \\
         & EBT & 14 & 53 \\
         & ACH & 30 & 111 \\
         & TLN & 1 & 108 \\
         \bottomrule
    \end{tabular}    
    }
\end{table}

\section{Extended Quantitative Results}
The impact of the meta-parameters \textit{Scenario}, \textit{Lighting} and \textit{Location} on model performance is displayed in \cref{tab:performance_impact}. Additional experiments indicate that the performance difference between YOLO11 models trained at input resolutions of $640^2$ and $1024^2$ pixels is generally negligible (\cref{tab:mAP_per_platform_1024}), with the sole exception of the OpenVINO INT8 model executed on the Raspberry Pi 5.

\begin{table}
    \caption{Impact of the meta-parameters \textit{Scenario}, \textit{Lighting} and \textit{Location} on model performance, measured on the test samples of the Core-Manual benchmark. Scores are image-averaged mAP\textsuperscript{50-95} (\%), i.e. mAP is computed per image and averaged over the subset, and Delta gives the difference to the overall image-averaged score of 69.8\% in percentage points. This metric is not directly comparable to the pooled evaluation in Tab. 3, which ranks all detections of the test set jointly.}   
    \label{tab:performance_impact}
    \centering
    \resizebox{0.45\linewidth}{!}{
    \begin{tabular}{@{}l@{\quad}l@{\quad}r@{\quad}r@{}}
        \toprule
        \textbf{Parameter} & \textbf{State} & \textbf{Score} & \textbf{Delta} \\
        \midrule
        \multirow{4}{*}{\textbf{Scenario}} & Front & 60.0 & -9.8 \\
         & Side & 69.9 & 0.0 \\
         & Back & 68.6 & -1.2 \\
         & Ground & 78.8 & 9.0 \\
         \midrule
        \multirow{3}{*}{\textbf{Lighting}} & Sunny & 68.3 & -1.5 \\
         & Mixed & 63.9 & -5.9 \\
         & Diffuse & 71.4 & 1.6 \\
         \midrule
        \multirow{4}{*}{\textbf{Location}} & KOR & 70.5 & 0.7 \\
         & EBT & 68.4 & -1.4 \\
         & ACH & 62.0 & -7.8 \\
         & TLN & 78.8 & 9.0 \\
         \bottomrule
     \end{tabular} 
     }
\end{table}

\begin{table}
    \caption{Detection performance (\% mAP\textsuperscript{50--95}) of YOLO11-Large models at an input resolution of $1024^2$ pixels across different deployment platforms, model formats (PyTorch, ONNX and OpenVINO) and numerical precisions (FP32, FP16 and INT8). An \textit{x} indicates unsupported combinations, a colon denotes identical values across a column. Hailo accelerators were evaluated exclusively as Raspberry Pi 5 HATs.}
    \label{tab:mAP_per_platform_1024}
    \centering
    \resizebox{0.85\linewidth}{!}{
    \begin{tabular}{@{}lc@{\quad}c@{\hspace{0.5em}}c@{\quad}c@{\hspace{0.5em}}c@{\hspace{0.5em}}c@{\quad}c@{\hspace{0.5em}}c@{}}
        \toprule
        & \multicolumn{1}{c@{\quad}}{\textbf{PyTorch}} & \multicolumn{2}{c@{\quad}}{\textbf{ONNX}} & \multicolumn{3}{c@{\quad}}{\textbf{OpenVINO}} & \multicolumn{2}{c}{\textbf{Hailo}} \\
        \textbf{Platform} & FP32 & FP32 & FP16 & FP32 & FP16 & INT8 & 8 & 8L \\
        \midrule
        Raspberry Pi 5 & : & : & 54.2 & : & 54.3 & 16.3 & 55.7 & 55.1 \\
        Intel W-2125 & : & : & 54.3 & : & 54.3 & 55.1 & x & x \\
        AMD 7430U & 55.3 & 54.3 & 54.3 & 54.3 & 54.3 & 55.2 & x & x \\
        AMD 9950X3D & : & : & 54.3 & : & 54.1 & 55.4 & x & x \\
        NVIDIA RTX 3090 & : & : & 54.3 & x & x & x & x & x \\
        \bottomrule
    \end{tabular}
    }
\end{table}

\begin{table}
    \caption{Inference time (ms) of YOLO11-Large models at an input resolution of $1024^2$ pixels across different deployment platforms, model formats (PyTorch, ONNX and OpenVINO) and numerical precisions (FP32, FP16 and INT8). An \textit{x} indicates unsupported combinations. Hailo accelerators were evaluated exclusively as Raspberry Pi 5 HATs.}
    \label{tab:ms_per_platform_640}
    \centering
    \resizebox{0.85\linewidth}{!}{
    \begin{tabular}{@{}l@{\quad}r@{\qquad}r@{\hspace{0.5em}}r@{\quad}r@{\hspace{0.5em}}r@{\hspace{0.5em}}r@{\quad}r@{\hspace{0.5em}}r@{}}
        \toprule
        & \multicolumn{1}{c@{\quad}}{\textbf{PyTorch}} & \multicolumn{2}{c@{\quad}}{\textbf{ONNX}} & \multicolumn{3}{c@{\quad}}{\textbf{OpenVINO}} & \multicolumn{2}{c}{\textbf{Hailo}} \\
        \textbf{Platform} & FP32 & FP32 & FP16 & FP32 & FP16 & INT8 & 8 & 8L \\
        \midrule
        Raspberry Pi 5 & 1,961 & 1,299 & 1,299 & 568 & 529 & 1,136 & 47 & 70 \\
        Intel W-2125 & 435 & 345 & 238 & 172 & 185 & 114 & x & x \\
        AMD 7430U & 476 & 345 & 345 & 250 & 244 & 118 & x & x \\
        AMD 9950X3D & 91 & 128 & 128 & 19 & 19 & 12 & x & x \\
        NVIDIA RTX 3090 & 11 & 9 & 9 & x & x & x & x & x \\
        \bottomrule
    \end{tabular}
    }
\end{table}

\end{document}